\documentclass[10pt,twocolumn,letterpaper]{article}

\usepackage{cvpr}
\usepackage{times}
\usepackage{epsfig}
\usepackage{graphicx}
\usepackage{amsmath}
\usepackage{amssymb}
\usepackage{subfigure}
\usepackage{enumitem}
\makeatletter
\newcommand{\printfnsymbol}[1]{  \textsuperscript{\@fnsymbol{#1}}}

\usepackage[breaklinks=true,bookmarks=false]{hyperref}

\cvprfinalcopy 
\def\cvprPaperID{****} 

\ifcvprfinal\fi
\begin{document}

\title{Attention-guided Network for Ghost-free High Dynamic Range Imaging}
\author{Qingsen Yan$^{1,2\dag}$, Dong Gong$^{2\dag}$, Qinfeng Shi$^{2}$, \\Anton van den Hengel$^{2}$, Chunhua Shen$^{2}$, Ian Reid$^{2}$, Yanning Zhang$^{1}$\thanks{$^\dag$The first two authors contributed equally to this work.
This work was partially supported by NSFC (61871328), ARC (DP140102270, DP160100703). Q. Yan and Y. Zhang were partially supported by National Engineering Laboratory for Integrated Aero-Space-Ground-Ocean Big Data Application Technology. Q. Yan was supported by a scholarship from CSC.}\\
$^1$School of Computer Science and Engineering, Northwestern Polytechnical University, China\\
$^2$The University of Adelaide, Australia\\
{\tt\small \url{https://donggong1.github.io/ahdr}}
}

\maketitle

\begin{abstract}
Ghosting artifacts caused by moving objects or misalignments is a key challenge in high dynamic range (HDR) imaging for dynamic scenes. Previous methods first register the input low dynamic range (LDR) images using optical flow before merging them, which are error-prone and cause ghosts in results. A very recent work tries to bypass optical flows via a deep network with skip-connections, however, which still suffers from ghosting artifacts for severe movement. To avoid the ghosting from the source, we propose a novel attention-guided end-to-end deep neural network (AHDRNet) to produce high-quality ghost-free HDR images. Unlike previous methods directly stacking the LDR images or features for merging, we use attention modules to guide the merging according to the reference image. The attention modules automatically suppress undesired components caused by misalignments and saturation and enhance desirable fine details in the non-reference images. In addition to the attention model, we use dilated residual dense block (DRDB) to make full use of the hierarchical features and increase the receptive field for hallucinating the missing details. The proposed AHDRNet is a non-flow-based method, which can also avoid the artifacts generated by optical-flow estimation error. Experiments on different datasets show that the proposed AHDRNet can achieve state-of-the-art quantitative and qualitative results.
\end{abstract}

\section{Introduction}
The dynamic range of natural luminance values varies over several orders of magnitude. However, most digital photography sensors can only measure a limited fraction of this range. The resulting low dynamic range (LDR) images thus often have over or underexposed regions and don't reflect the human ability to see details in both bright and dark areas of a scene.
High dynamic range (HDR) imaging has been developed to compensate for these limitations, and ideally aims to generate a single image that represents a broad range of illuminations.
\begin{figure}[!t] 
\def \wid{8cm} 
\centering
\includegraphics[width=\wid]{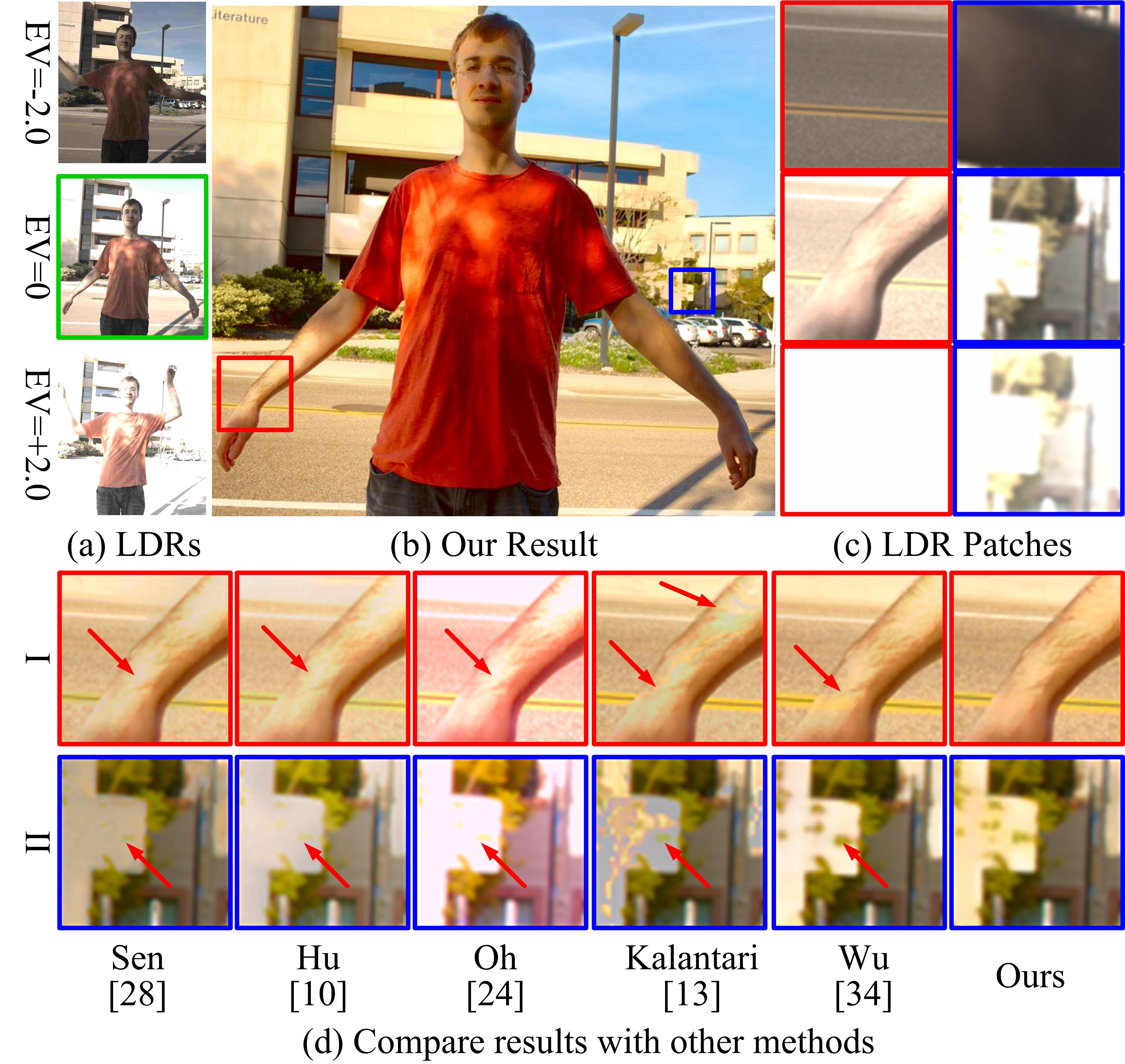}
\caption{
LDR images with different exposures are shown in (a), and our result after tonemapping is shown in (b).
The areas of the images that exhibit both large-scale movement and saturation are displayed in (c).
The proposed AHDRNet generates an HDR image with less ghosting artifacts and more details in saturated regions (See zoomed-in patches in (d)).
}
\label{Fig_1}
\end{figure}
\par
Some specialized hardware devices \cite{Nayar2002High,Tublin2005} have been proposed to produce HDR images directly, but they are usually too expensive to be widely adopted.
As a result, computational HDR imaging methods have drawn more attention. The most common strategy is to take a series of LDR images at different exposures and then merge them into an HDR image \cite{Debevec1997Recovering,Picard95onbeing,Granados2010Optimal,Reinhard2005High,YAN2017}.  In multiple exposure methods, one of the LDR images is usually considered as the reference image (shown with the green border in Figure \ref{Fig_1} (a)). Although these methods often generate high-quality HDR results when the scene and camera are completely static, they will suffer from significantly ghosting and blurring artifacts when there is motion between the input images.

\par
Global image misalignments can be compensated for using homographies \cite{szpak2014sampson,szpak2015robust,Wu_2018_ECCV}. However, the ghosting artifacts caused by moving objects and the missing details due to saturation are complex to overcome.
To tackle the ghosting issue, some methods first carry out a more detailed alignment of the LDR images before merging \cite{Grosch2006,Jacobs2008,SRIKANTHA2012survey}. A variety of alignment procedures have been applied (\eg optical flow \cite{Kalantari2017Deep,KangSig2003,Zimmer2011Freehand}), but they still suffer from the artifacts due to the estimation error.
To avoid this alignment error, some methods \cite{Oh2015pami,Raman2011} proposed to reject the misaligned moving components as outliers directly. However, pixel-accurate identification of moving objects is difficult to achieve robustly, particularly when relying on simple pixel level characteristics (\eg pixel color \cite{Raman2011}).

\par
Inspired by the successes of the deep neural networks (DNNs) in many image restoration tasks \cite{ledig2017photo,Gong2016From,yang2018seeing,gong2018learning}, some deep learning-based approaches \cite{Kalantari2017Deep,Wu_2018_ECCV,YANwacv2019} have been proposed recently to improve the HDR image composition process. In \cite{Kalantari2017Deep}, a DNN is proposed to merge the LDR images after an optical flow based alignment process. However, the DNN cannot handle the distortions caused by the inevitable optical flow estimation error (See Kalantari \emph{et al.}'s method in Figure \ref{Fig_1} (d)). In \cite{Wu_2018_ECCV}, the HDR imaging task is treated as an image translation problem. Although the model can produce satisfactory results in some examples, it still suffers from ghosting artifacts when there are large-scale movements between the images. The DNN-based methods can hallucinate some details in regions with saturation, but the existing methods cannot handle large areas of saturation, particularly when there is also occlusion.

\par
\begin{figure*}[t!]
\centering
\def \wid{16.5cm}
\includegraphics[width=\wid]{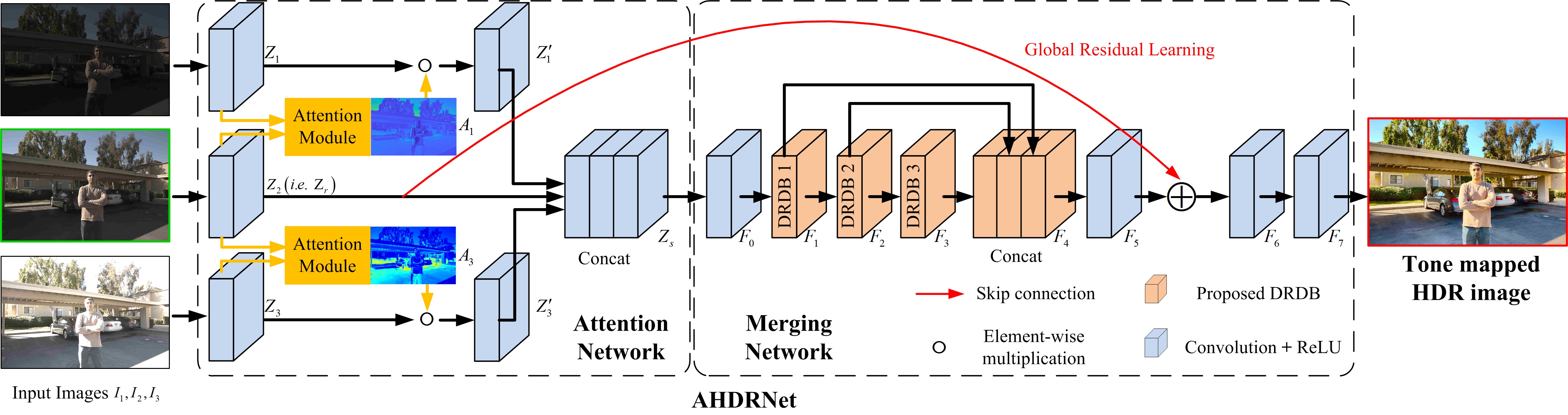}
\caption{The architecture of the proposed AHDRNet. The network consists of an \emph{attention network} for feature extraction and a \emph{merging network} for predicting the HDR image.
The attention module is used to exclude the harmful components caused by misalignment and saturation or highlight the useful details.
The merging network is constructed based on a series of dilated residual dense blocks (DRDBs).
The global residual skip connection is used to boost the training.
The final HDR result is obtained by tonemapping. All the feature maps have 64 channels, and the kernel size is 3.
The visualized map is a presentation of the averaged attention feature $A_i$.
}
\label{Fig_flowchart}
\end{figure*}

We propose an attention-guided deep neural network (AHDRNet) for HDR imaging (See Figure \ref{Fig_flowchart}).
The neural network learns the relationships between input LDR images and HDR output.
Previous methods \cite{Kalantari2017Deep,Wu_2018_ECCV} take stacked LDR images, or LDR image feature maps, as the input to the merging process, which mixes the misaligned image components at an early stage of the network, making it difficult to obtain ghost-free HDR results. Considering that ghosting is primarily an artifact of object motion and misalignments \cite{Kalantari2017Deep}, we propose the learnable attention modules to guide the merging process.
The attention modules generate soft attention maps to evaluate the importance of different image regions for obtaining the required HDR image. They are expected to highlight the features complementary to the reference image and exclude regions with motion and severe saturation. The LDR image features with attention guidance are then fed to the merging network to generate the HDR image.
We construct the merging network using dilated residual dense blocks (DRDBs), which are achieved by employing the dilated convolution layers in the residual dense block (RDB) proposed in \cite{zhang2018RDB}. The RDBs help to make full use of information from different convolutional layers, thus preserving more details from the input LDR images. The dilated convolutions enlarge the receptive fields, helping to recover the details contaminated by saturation and moving objects.
The main contributions of the paper can be summarized as:

\begin{itemize}[itemsep=-1pt,topsep=-2pt]
\item We propose a new attention-guided network for ghost-free HDR imaging.
It has all of the benefits of a neural network model, and overcomes one of the primary problems in HDR imaging is that it is robust to large misalignments of image pixels and saturation.

\item We propose a network based on dilated residual dense blocks to merge the attention guided feature maps from LDR images. The dilated residual dense blocks can simultaneously preserve the image details and enlarge the receptive fields, allowing the network to hallucinate the contents in saturated regions and produce HDR images with rich details.

\item Extensive experiments on different datasets validate the superiority of the proposed AHDRNet. We also conduct ablation studies to quantify the roles of different components in our model.

\end{itemize}

\section{Related Work}
\label{sec2}

The primary relevant works are as follows.

\noindent \textbf{Methods relying on pixel rejection}~ These approaches label each pixel as belonging to a static region or a moving object based on the assumption that the images are globally registered.
Grosch \cite{Grosch2006} defined an error map that uses the color difference of inputs to get the ghost-free HDR image.
Jacobs \emph{et al.} \cite{Jacobs2008} detected ghost regions based on a weighted variance measure.
Pece and Kautz \cite{Pece2010Bitmap} computed the median threshold bitmap for input images to detect motion regions.
Heo \emph{et al.} \cite{Heo2011ACCV} roughly detected motion regions by joint probability densities and these regions are refined using energy minimization based on graph-cuts methods.
Zhang and Cham \cite{GradientHDR2012} proposed quality measures based on image gradients to generate a weighting map over the inputs.
Rank minimization \cite{Lee2014Ghost,Oh2015pami} has also been used to detect motion regions and reconstruct HDR images.
Even it is achieved to the required pixel accuracy, rejecting pixels reduces the information available to reconstruct the HDR image, which often leads to missing details (See Oh's method \cite{Oh2015pami} in Figure \ref{Fig_1}).

\noindent \textbf{Methods relying on registration}~ These approaches reconstruct each HDR region by searching for the best matching region in LDR images.
This is achieved using pixel (optical flow methods) or patch (patch-based methods) based dense correspondences.
Bogoni \cite{Bogoni2000} estimated motion vectors using optical flow and used parameters to warp pixels in the exposures.
Kang \emph{et al.} \cite{KangSig2003} transformed intensities of LDR images to the luminance domain using exposure time information and
computed the optical flow to find corresponding pixels among the LDR images.
Sen \emph{et al.} \cite{Sen2012} proposed a patch-based energy minimization approach that integrates alignment and HDR reconstruction in a joint optimization.
Hu \emph{et al.} \cite{Hu2013deghosting} optimized image alignment based on brightness and gradient consistencies on the transformed domain.
Hafner \emph{et al.} \cite{Hafner2014Simultaneous} proposed an energy-minimization approach which simultaneously calculates HDR irradiance and displacement fields.
This approach improves robustness, but fails for large motions, doesn't learn by examples, and makes no attempt to compensate for saturation.

\noindent \textbf{Deep learning based methods}~
Many deep learning approaches \cite{Eilertsen2017HDR,Kalantari2017Deep,Wu_2018_ECCV} have been developed.
Eilertsen \emph{et al.} \cite{Eilertsen2017HDR} proposed a deep autoencoder network to predict HDR values from one image.
Endo \cite{Endo2017Deep} synthesized multiple LDR images from one LDR image with the deep-learning-based approach, then reconstructed an HDR image by merging them.
Kalantari \emph{et al.} \cite{Kalantari2017Deep} used optical flow to align the input images to the reference image, then employed a convolutional neural network to obtain the HDR image.
Wu \emph{et al.} \cite{Wu_2018_ECCV} proposed a network that can learn to translate multiple LDR images into a ghost-free HDR image.
These methods have the advantage that they can exploit information extracted from training data to identify and compensate for image regions that do not meet the assumptions underlying the HDR process. Each method addresses an important issue, but none has the flexibility and robustness that the proposed attention-based approach enables (See Figure~\ref{Fig_1}).

\noindent \textbf{Attention mechanisms in deep learning methods}~
Attention has shown to be a pivotal development in deep learning and has been used in many computer vision applications.
Lu \emph{et al.} \cite{Lu2017Adaptive} proposed a novel adaptive attention model with a visual sentinel for image captioning.
Fan \emph{et al.} \cite{Fan2018CVPRatt} stacked latent attention for multiple multimodal reasoning tasks.
Zhao \emph{et al.} \cite{Zhao2017att} proposed a diversified visual attention network to address the problem of fine-grained object classification.
Each has achieved the hitherto impossible performance and robustness by allowing models to focus on only the relevant information.

\section{Attention-guided Network for HDR Imaging}
\label{sec3}
Given a series of LDR images of a dynamic scene $(I_1, I_2, ..., I_k)$ with different exposures, the target of HDR imaging is to recover an HDR image $H$ aligned to a prescribed reference image $I_r$ (selected from the input LDR images). All of the images $I_i$ and $H$ are RGB images with three channels.
Following the settings in \cite{Kalantari2017Deep,Wu_2018_ECCV}, we use three LDR images $(I_1, I_2, I_3)$ (sorted by their exposure lengths), \ie $k=3$, and let the middle exposure image $I_2$ be the reference image.

\par
Before feeding the LDR images to the network, we first map the input LDR images $\{I_i\}$ to the HDR domain relying on gamma correction \cite{Kalantari2017Deep,Wu_2018_ECCV} to generate a corresponding set of $\{H_i\}$:
\begin{equation}
    H_i = I_i^{\gamma} / t_i, ~\forall i=1,2,3,
\end{equation}
where ${\gamma}>1$ denotes the gamma correction parameter and $t_i$ denotes the exposure time of the image $I_i$. We set $\gamma=2.2$ in this work.
As suggested in \cite{Kalantari2017Deep}, we concatenate images $I_i$ and $H_i$ along the channel dimension to obtain the 6-channel tensors $X_i=[I_i, H_i], i=1,2,3$ as the input of the network. Intuitively, the LDR images $L_i$ help to identify the noisy and saturated regions, while the $H_i$ facilitate the detection of the alignments\cite{Kalantari2017Deep}.
Given $(X_1, X_2, X_3)$ as input, the proposed AHDRNet obtains the HDR image by
\begin{equation}
H = f(X_1, X_2, X_3; \theta),
\end{equation}
where $f(\cdot)$ denotes the proposed HDR network, and $\theta$ is the network parameters.
The attention mechanism works as part of the end-to-end AHDRNet network $f(\cdot)$.
Note that the input images of the proposed model can be the original images without any alignment preprocessing.

\subsection{Overview of the AHDRNet Architecture}
Unlike the previous methods \cite{Kalantari2017Deep,Wu_2018_ECCV} that stack the input images $X_i$ or the extracted feature maps in the early stage of the network for merging, the proposed AHDRNet obtains the attention maps by comparing the encoded image features and then merges features with the guidance of the attention maps.
As shown in Figure \ref{Fig_flowchart}, the AHDRNet consists of two major subnetworks, \ie the \emph{attention network} (for feature extraction) and the \emph{merging network} (for HDR image estimation).

\par
The \emph{attention network} first separately extracts features from each LDR image relying on the corresponding convolutional encoders.
Then, we apply specific attention maps on the \emph{non-reference images} to identify the beneficial features. The attention maps are obtained via the attention modules according to the feature maps from the reference image and each non-reference image.
Considering that the target of the model is to generate the HDR image with the scene consistent to the reference image, the motivation of applying attention on the non-reference images is to identify the misaligned components before merging the features for alleviating the ghosting artifacts.

\par
The \emph{merging network} takes the features extracted with the attention guidance as input and estimates the HDR image relying on a series of dilated residual dense blocks (DRDBs) and the global residual learning (GRL) strategy. The DRDBs and GRL help to utilize the image features effectively and obtain the HDR image with plausible details. The merging network fuses the features from the LDR images and hallucinates the details in the regions contaminated by the saturation and misaligned moving objects.

\subsection{Attention Network for Feature Extraction}
Given three 6-channel input images $X_i, i=1,2,3$ corresponding to the three LDR images, the attention network first uses a shared encoding layer to extract feature maps $Z_i, i=1,2,3$ with 64 channels from three inputs.
For clarity, we define notations $X_r$ and $Z_r$ to indicate $X_2$ and $Z_2$ corresponding to the reference LDR image in some special context. As shown in Figure \ref{Fig_flowchart}, to obtain the attention maps for the non-reference images, we feed the features $Z_i, i=1,3$ of the non-reference images to the convolutional \emph{attention module} $a_i(\cdot), i=1,3$ along with the reference image feature map $Z_r$, and then obtain the attention maps $A_i$ for the non-reference images:
\begin{equation}
A_i = a_i(Z_i, Z_r), ~i=1, 3.
\label{eq:att_map}
\end{equation}
$A_i$ has the same size as $Z_i$. The values in $A_i$ are in the range $[0,1]$.
Details of the \emph{attention modules} are provided below.
The predicted attention maps are used to attend the features of the non-reference images via:
\begin{equation}
Z'_i = A_i \circ Z_i, ~i=1, 3,
\label{eq:attention_opt}
\end{equation}
where $\circ$ denotes the point-wise multiplication and $Z'_i$ denotes the feature maps with attention guidance.

\par
Instead of stacking the original feature maps $Z_i$'s for HDR merging, we stack the reference feature map $Z_r$ (\ie $Z_2$) and the features of the non-reference images $Z'_1$ and $Z'_3$ for merging. The attention network thus obtains a stack of features with the guidance of the reference as $Z_s$
\begin{equation}
Z_s=\text{Concat}(Z'_1, Z_2, Z'_3),
\end{equation}
where $\text{Concat}(\cdot)$ denotes the concatenation operation. $Z_s$ will be used as the input of the merging network.

\begin{figure}[t]
\def \wid{8cm}
\centering
\includegraphics[width=\wid]{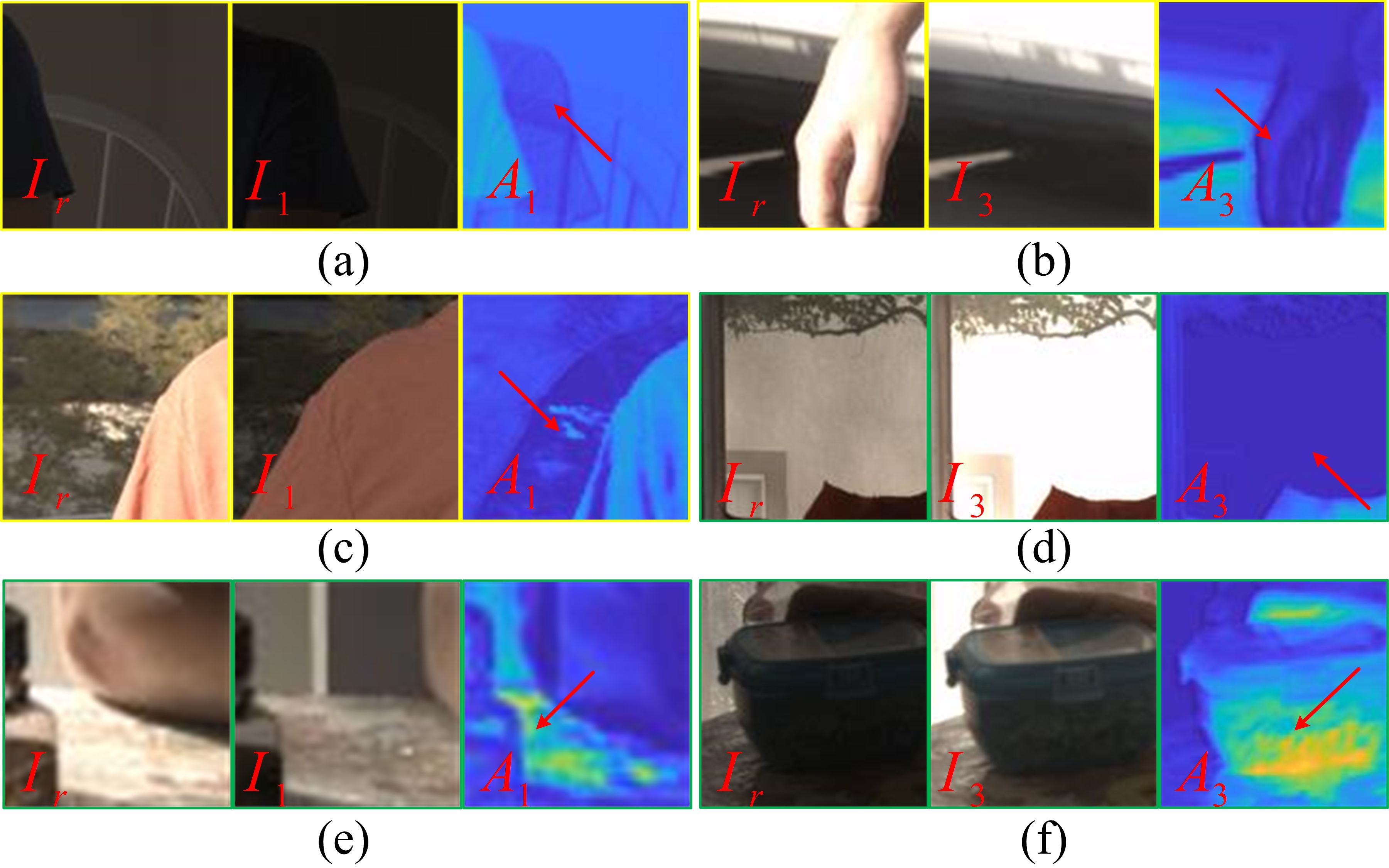}
\caption{Example image patches and the corresponding attention maps. In (a)-(f), from left to right: the reference image, one non-reference image, and the attention map applied on the non-reference image.
(a), (b) and (c) display attention maps for the significantly misaligned regions. (d), (e) and (f) show the attention maps can highlight useful regions.}
\label{Fig_attmap_small}
\vspace{-0.5cm}
\end{figure}

\par
Since the HDR imaging process centers on the reference image, the attention maps are predicted and applied according to the reference. As shown in Figure \ref{Fig_attmap_small}, the attention maps can suppress the misaligned (See (a) (b) and (c)) and saturated regions (See (d)) in the non-reference images, which avoids the harmful features getting into the merging process and thus alleviates the ghosts from the source.
When some regions in the reference are saturated (See (e)) or noisy (See (f)), the attention maps can also highlight useful features in the non-reference images.
More studies in Section \ref{sec_ablation} further prove the effectiveness of the proposed attention mechanism in HDR imaging.

\begin{figure}[t]
\def \wid{8.5cm}
\centering
\includegraphics[width=\wid]{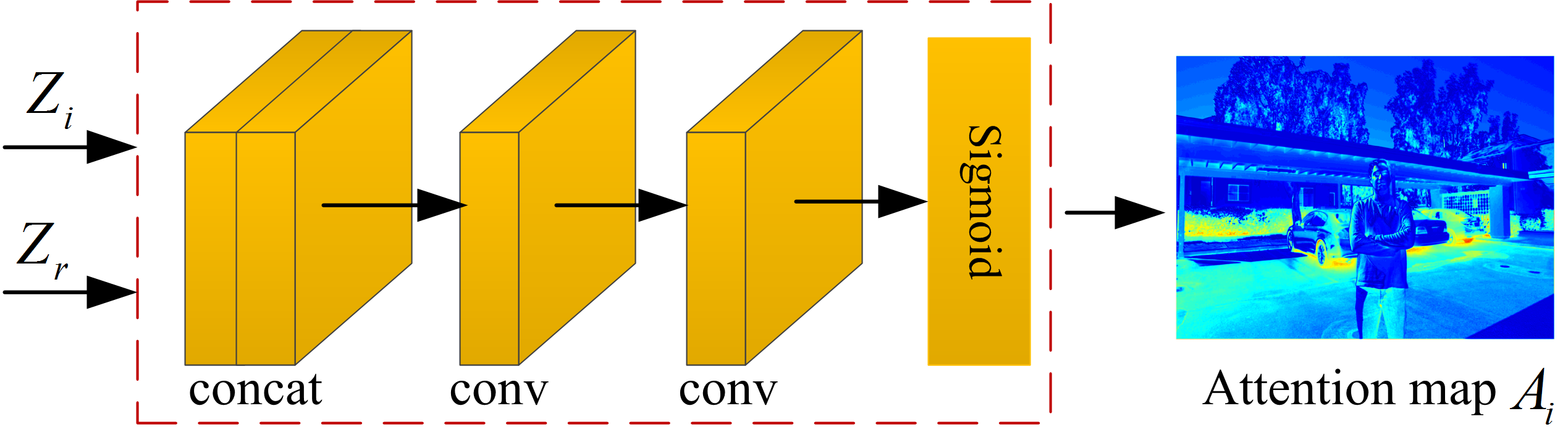}
\caption{The attention unit first concatenates the two inputs and then obtains attention maps via two Conv layers, which restricts the output in [0,1] using a sigmoid activation. }
\label{Fig_attmap}
\end{figure}

\noindent \textbf{Attention module}~ The attention modules $a_i(\cdot), i=1,3$ in Eq. \eqref{eq:att_map} are two small CNNs. The structure of the attention modules is shown in Figure \ref{Fig_attmap}. The attention module first concatenates the input feature maps $Z_i$ and $Z_r$ and obtains the attention map after two convolution (Conv) layers. Each Conv layer applies 64 $3\times 3$ layers. The two Conv layers are followed by a ReLU activation and a sigmoid activation, respectively.
As a result, the attention module can obtain the 64-channel attention map $A_i$ with values in the range $[0,1]$.

\subsection{Merging Network for HDR Image Estimation}
\label{sec:merging_network}
The merging network takes the stacked feature map $Z_s$ and the reference image feature map $Z_r$ as input.
In the design of the merging network, we take account of the characteristics of the HDR imaging problem and use the basic structure of the residual dense network in \cite{zhang2018RDB} as the reference.
As shown in Figure \ref{Fig_flowchart}, the network consists of several convolution layers, dilated residual dense blocks and several skip connections. The generated feature maps at different layers are noted as $F_j, j=0,1,...,7$.

\par
Given the stacked feature $Z_s$, the merging network first obtains a 64-channel feature map after a Conv layer, and then feed it into three DRDBs, which results in three corresponding feature maps $F_1$, $F_2$ and $F_3$. Instead of using the RDB proposed in \cite{zhang2018RDB}, we proposed to use the RDBs with dilated convolution (DRDB) for HDR imaging. The details of DRDB can be found in the following.
By applying $3\times 3$ Conv on the concatenated feature map $F_4$, we generate the merged and transferred feature map $F_5$.

\par
\noindent \textbf{Global residual learning with the reference features}~
Before reconstructing the HDR image from $F_5$, inspired by the super-resolution methods \cite{ledig2017photo,zhang2018RDB}, we apply a global residual learning strategy to obtain feature maps by
\begin{equation}
F_6 = F_5 + Z_r,
\label{eq:global_residual_learning}
\end{equation}
where $Z_r$ is the shallow feature map of the reference image. The merging network thus tends to learn the residual features. In the proposed AHDRNet, we have the shallow feature map $Z_r$ containing the pure information from the reference image. We thus apply the global residual learning with the reference feature maps.
We consider that the feature map $F_6$ contains enough information to reconstruct the HDR image.
Empirical studies in Section \ref{sec_ablation} show the effectiveness of the global residual learning strategy.

\begin{figure}[t]
\def \wid{8cm}
\centering
\includegraphics[width=\wid]{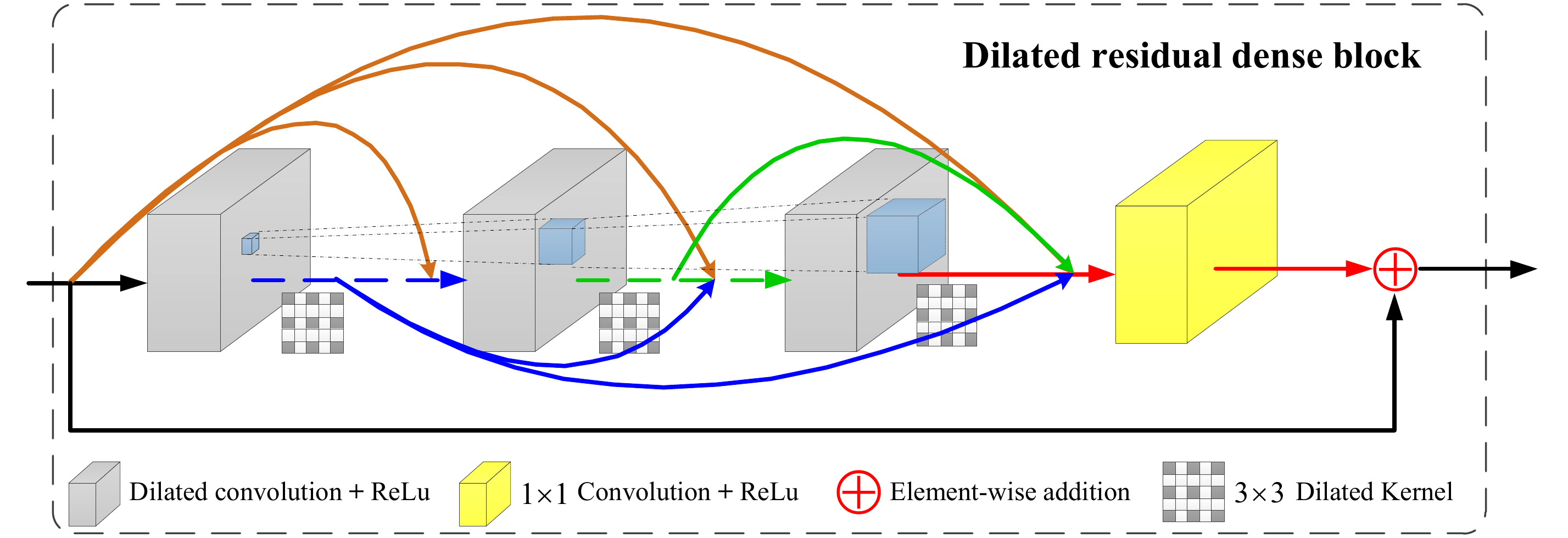}
\caption{Illustration of dilated residual dense block structure with three convolution layers. We adopt a residual dense block \cite{zhang2018RDB} as its backbone and each convolution layer can be substituted by dilated convolution.
By using dilated residual dense blocks, the receptive field at each block is expanded.}
\label{Fig_DRDB}
\end{figure}

\par
After two convolution layers (followed by activations), we estimate the HDR image $\widehat{H}$ in the HDR domain. The final HDR image is displayed via the tonemapping operation (See Section \ref{sec:training_loss}).

\noindent \textbf{Dilated residual dense block}~
Since the reconstruction of some local areas of the HDR images cannot get enough information from the LDR images due to the occlusion of moving objects and saturation,
the merging network requires larger receptive field for hallucinating details. We thus apply the 2-dilated convolutions \cite{yu2015multi} in the residual dense block (RDB) \cite{zhang2018RDB}.
As shown in Figure \ref{Fig_DRDB}, the proposed dilated residual dense block (DRDB) consists of a series of Conv layers followed by ReLU activations and dense concatenation based skip-connections.
Each Conv layer takes the concatenation of all the feature maps from previous layers as input.
In contrast to the dense block proposed in \cite{huang2017densely}, the RDB and DRDB apply a local residual skip-connection between the input and output of a block.
More details of the RDB can be found in \cite{zhang2018RDB}.
In our implementation, we use 6 Conv layers in each DRDB.
The empirical ablations studies in Section \ref{sec_ablation} show the effectiveness of the DRDBs.

\subsection{Training Loss}
\label{sec:training_loss}
As described in Section \ref{sec:merging_network}, the proposed AHDRNet predicts the HDR image $\widehat{H}$ in the HDR domain. Since the HDR images are usually displayed after tonemapping, training the network on the tonemapped images is more effective than training directly in the HDR domain \cite{Kalantari2017Deep}.
Given an HDR image ${H}$ in HDR domain, we compress the range of the image using $\mu$-law:
\begin{equation}
\mathcal{T}(H)=\frac{\log(1+\mu H)}{\log(1+\mu)},
\label{eq:mu}
\end{equation}
where $\mu$ is a parameter defining the amount of compression and $\mathcal{T}(H)$ denotes the tonemapped image. In this work, we always keep $H$ in the range $[0,1]$ and set $\mu=5000$. The tonemapper in Eq. \ref{eq:mu} is differentiable, which is very suitable for training the network.

\par
In our method, we train the network by minimizing $\ell_1$-norm based distance between the tonemapped estimated and the ground truth HDR images. Our loss function is defined as:
\begin{equation}
\mathcal{L} = \| \mathcal{T}(\widehat{H}) - \mathcal{T}(H) \|_1.
\end{equation}
We also tested the $\ell_2$ loss used in previous work \cite{Kalantari2017Deep,Wu_2018_ECCV} and noticed that $\ell_1$ loss is more powerful for preserving details (See Section \ref{loss_compare}), which is consistent with the observation in \cite{zhao2017loss}.

\subsection{Implementation Details}
In our implementation, we apply 64 $3\times 3$ features in the Conv layers, which are followed by ReLU activations, if not specified otherwise. We set the stride size for all Conv layers as 1 and keep the feature map size using zero padding. We define the output layer to produce 3-channel images. The growth rate of all DRDBs is 32. The last Conv layer in each DRDB applies $1\times 1$ convolution to compress the feature maps.

\par
For training, we use Adam optimizer \cite{Kingma2014Adam} and set the batch size and learning rate as 8 and $1\times 10^{-5}$, receptively.
Given training images, we randomly crop the $256 \times 256$ patches for training.
All weights of the network are initialized using Xavier method \cite{Glorot2010Understanding}.
We implement our model using PyTorch \cite{paszke2017automatic}, which takes takes 0.32s to process a $1500\times 1000$ image with an NVIDIA GeForce 1080 Ti GPU.

\section{Experiments}
\label{sec4}

\subsection{Experimental Settings}
\label{sec_exp_setting}
\noindent \textbf{Training data}~
We train the AHDRNet on the HDR dataset \cite{Kalantari2017Deep} which includes 74 samples for training and 15 samples for testing.
For each sample, three different LDR images are captured with exposure biases of $\{-2, 0, +2\}$ or $\{-3, 0, +3\}$. Transformations on the cropped patches are applied as data augmentation to alleviate overfitting.

\noindent \textbf{Testing data}~
We test the proposed AHDRNet on the Kalantari's dataset \cite{Kalantari2017Deep} and the datasets without ground truth, such as Sen's \cite{Sen2012} and Tursun's \cite{Tursun2016data} datasets.

\noindent \textbf{Evaluation Metrics}
We conduct evaluations with four metrics as the following.
We compute the PSNR values for images after tonemapping using $\mu$-law (PSNR-$\mu$), Matlab function \textit{tonemap} (PSNR-M), and linear (PSNR-L) domains.
We also conducted a quantitative evaluation by computing the HDR-VDP-2 \cite{Mantiuk2011HDR}.

\subsection{Ablation Studies}

\subsubsection{Study on the Model Architecture}
\label{sec_ablation}
\par
We investigate the architecture of AHDRNet and validate the importance of different individual components in the whole AHDRNet. We achieve this ablation study by comparing the proposed AHDRNet and the following variants of AHDRNet:

\begin{itemize}[itemsep=-1pt,topsep=-2pt]
\item \textbf{AHDRNet}. The full model of the AHDRNet.
\item \textbf{DRDB-Net} (\ie AHDRNet w/o attention). We remove the attention module in this variant, in which the feature maps $Z_i$'s are directly stacked and fed to the merging network.
\item \textbf{A-RDB-Net} (\ie AHDRNet w/o dilation). We do not use dilated convolution in this variant of AHDRNet.
\item \textbf{RDB-Net} (\ie AHDRNet w/o attention and dilation). This variant of AHDRNet does not contain the attention operation and dilated convolution layers.
\item \textbf{RB-Net} (\ie AHDRNet w/o attention, dilation and densely connection). This baseline is a merging network based on the residual block (RB). We replace the DRDBs as the same number of RBs.
\item \textbf{Deep-RB-Net}. More RBs are used to approach the model compressibility of the RDB-Net.
\end{itemize}

\begin{table}
\small
\begin{center}
  \caption{Quantitative comparisons of different models. All scores are the average across all testing images.}
  \begin{tabular}{c|c|c|c|c}
  \hline
    &\footnotesize{PSNR-$\mu$} & \footnotesize{PSNR-M} & \footnotesize{PSNR-L} & \footnotesize{HDR-VDP-2}\\
     \hline
     \footnotesize{RB-Net}& 39.8648 &  28.3548 &  38.0044 &  60.1905\\
          \footnotesize{Deep-RB-Net}   & 41.1788  & 29.5414 &  38.9679 &  60.2724\\
          \footnotesize{RDB-Net}  & 41.2058  & 29.4335  & 38.9747  & 60.5107\\
          \footnotesize{DRDB-Net} & 42.7345  & 31.4169  & 39.7800  & 60.8740\\
          \footnotesize{A-RDB-Net} & 43.0536   &32.2025  & 40.5105  & 61.6362 \\
          \hline
        \footnotesize{w/o GRL} & 42.5313  & 32.9552 &  40.7558 &  62.2827\\
    \hline
     \footnotesize{AHDRNet}   & \textbf{43.6172}  & \textbf{33.0429 } & \textbf{41.0309 }&  \textbf{62.3044}\\
     \hline
\end{tabular}
\label{table:module}
\end{center}
\end{table}

\begin{figure}[tbp]
\def \wid{8.5cm}
\centering
\includegraphics[width=\wid]{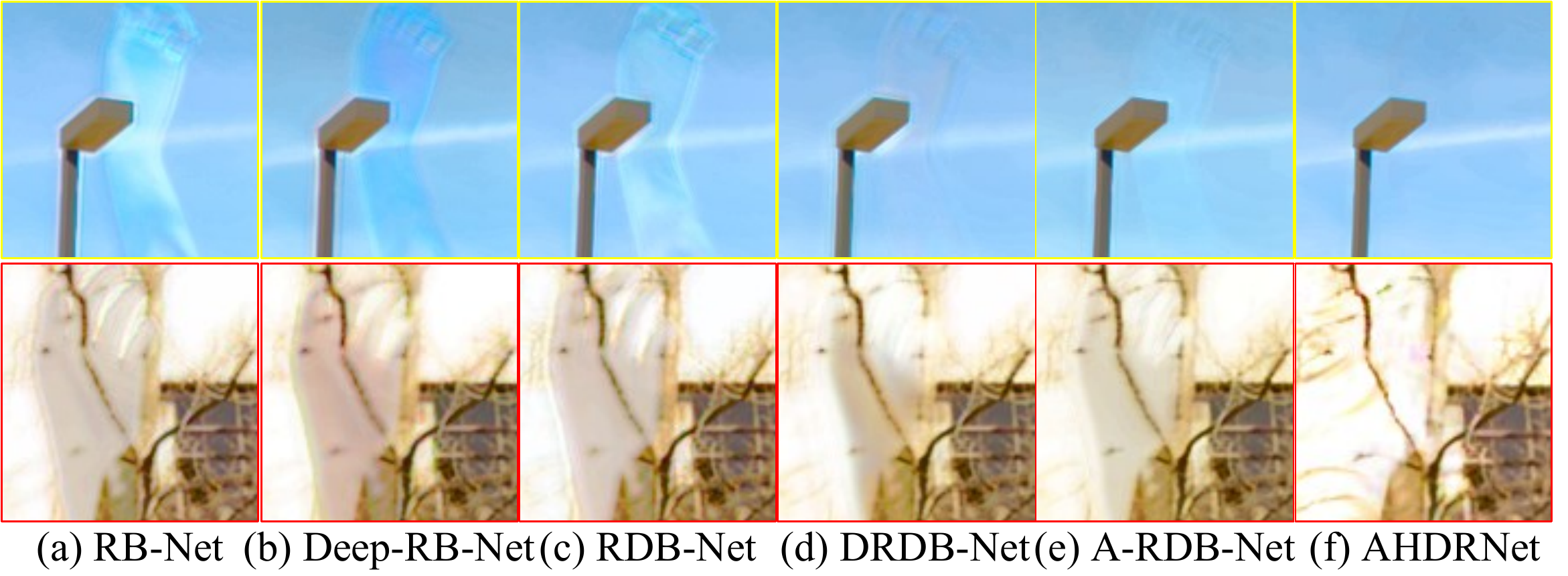}
        \caption{Visual results of AHDRNet and its baseline variants.}
\label{Fig_module}
\end{figure}

\noindent \textbf{Attention module.}~The attention module is a very effective mechanism for HDR image de-ghosting task.
As shown in Figure \ref{Fig_module}, compared with RDB-Net, A-RDB-Net can alleviate the ghosting artifacts due to the attention module.
A similar result can be observed with DRDB-Net and AHDRNet.
Although DRDB-Net can remove ghosting artifacts, it tends to generate artifacts in saturated regions (the bottom patch of Figure \ref{Fig_module}).
The proposed AHDRNet with attention module can eliminate ghosting artifacts while retaining the background information (See Figure \ref{Fig_attmap_small}).
As shown by the quantitative results in Table \ref{table:module},
AHDRNet and A-RDB-Net acquire a better improvement than the DRDB-Net and RDB-Net.

\noindent \textbf{Dilated residual dense blocks.}
Compared with DRDB-Net, the RB-Net results have visible ghosts (See Figure \ref{Fig_module} (a) and (d)).
Even the results of Deep-RB-Net that has more RBs cannot remove ghosting artifacts.
Hence, increasing the depth of the network is not a practical approach to enhance HDR image quality.
On the other hand, the DRDB-Net with the same network depth can capture more contents and alleviate ghosts.
The performance of DRDB-Net in Table \ref{table:module} is better than RB-Net and RDB-Net.

\noindent \textbf{Dilated convolution.}
To demonstrate the capability of dilated convolution, we compare the RDB-Net and DRDB-Net.
As displayed in Figure \ref{Fig_module} (c) and (d), the results of DRDB-Net alleviate ghosting artifacts compared with RDB-Net.
The results show that a larger receptive field is helpful to suppress the ghosting artifacts and hallucinate the missing details.
Furthermore, the proposed AHDRNet can completely remove ghosts.
The quantitative comparisons in Table \ref{table:module} show that
the models with dilated convolution can obtain high values on PSNR metrics.

\noindent \textbf{Global residual learning.}~
We also study the performance of global residual learning (GRL) strategy.
Quantitative comparisons of the results are shown in Table \ref{table:module}. Since GRL helps to transfer information from front layers, the model with GRL can bring better performance.

\subsubsection{Study on Training Loss Function}
\label{loss_compare}
\par
In this experiment, we compare the performances of our method with different loss functions.
Quantitative comparisons of the results are shown in Table \ref{tab:diff_loss}, which implies that the $\ell_1$ loss is more powerful for preserving details as discussed in \cite{zhao2017loss}.
We thus train our model using $\ell_1$ loss.

\begin{table}[htp]
\begin{center}
\small
  \caption{Quantitative comparisons of different loss functions.}
  \begin{tabular}{c|c|c|c|c}
  \hline
     \\     &\footnotesize{PSNR-$\mu$} & \footnotesize{PSNR-M} & \footnotesize{PSNR-L} & \footnotesize{HDR-VDP-2}\\
     \hline
     $\ell_2$ loss   & 43.0630  & 31.7921  & 40.6798 &  62.0169\\
     $\ell_1$  loss  & \textbf{43.6172 }  & \textbf{33.0429}  & \textbf{41.0309} &  \textbf{62.3044}\\
     \hline
\end{tabular}
\label{tab:diff_loss}
\end{center}
\end{table}

\begin{figure*}[htbp]
\def \wid{8.5cm}
\def \Lwid{0.495}
\centering
\begin{minipage}[h]{\Lwid\linewidth}
      \centering
      \centerline{\includegraphics[width=\wid]{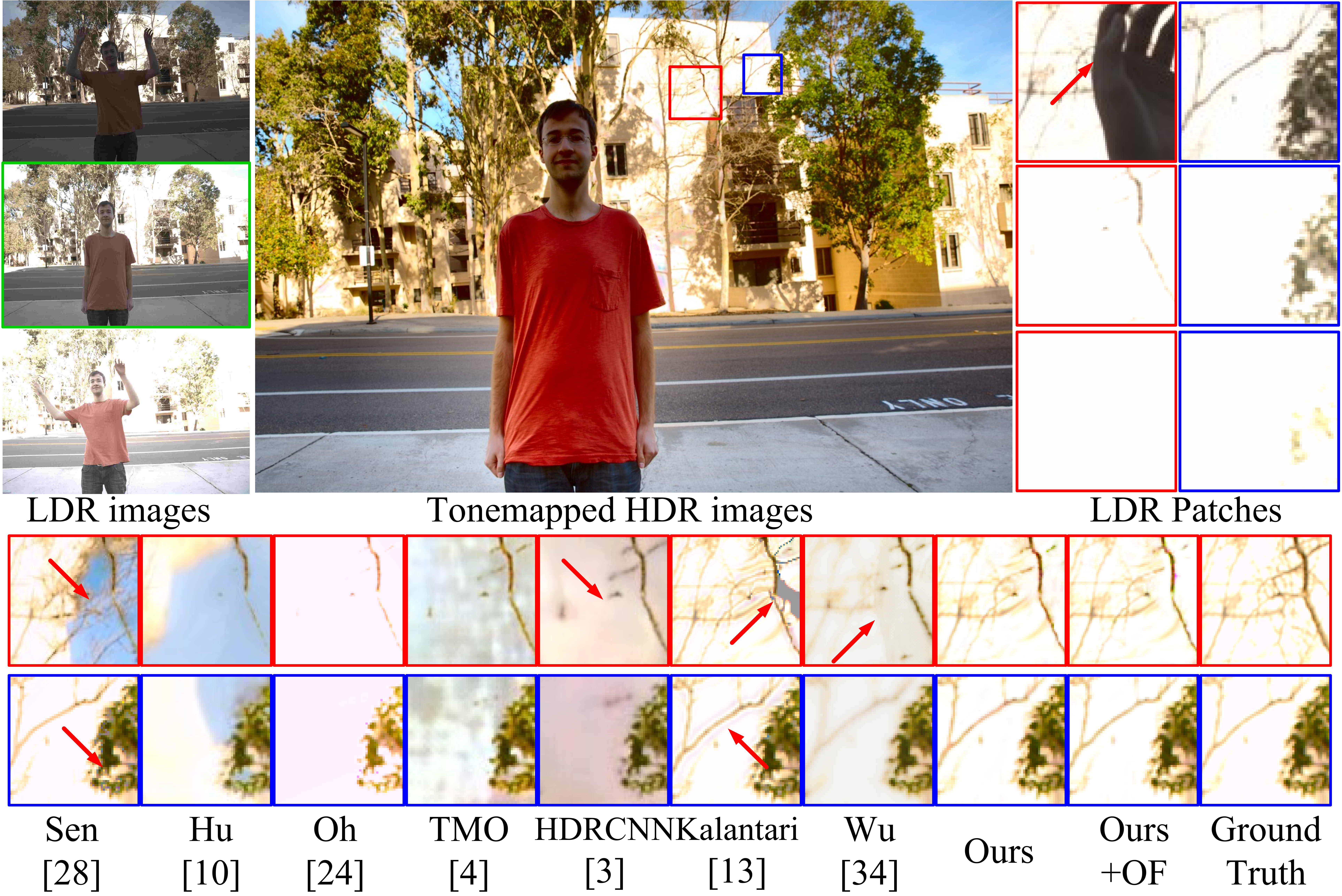}}
     \centerline{\begin{small}(a) Testing data (Building) in \cite{Kalantari2017Deep} \end{small}}\medskip
\end{minipage}
     \hfill
\begin{minipage}[h]{\Lwid\linewidth}
      \centering
      \centerline{\includegraphics[width=\wid]{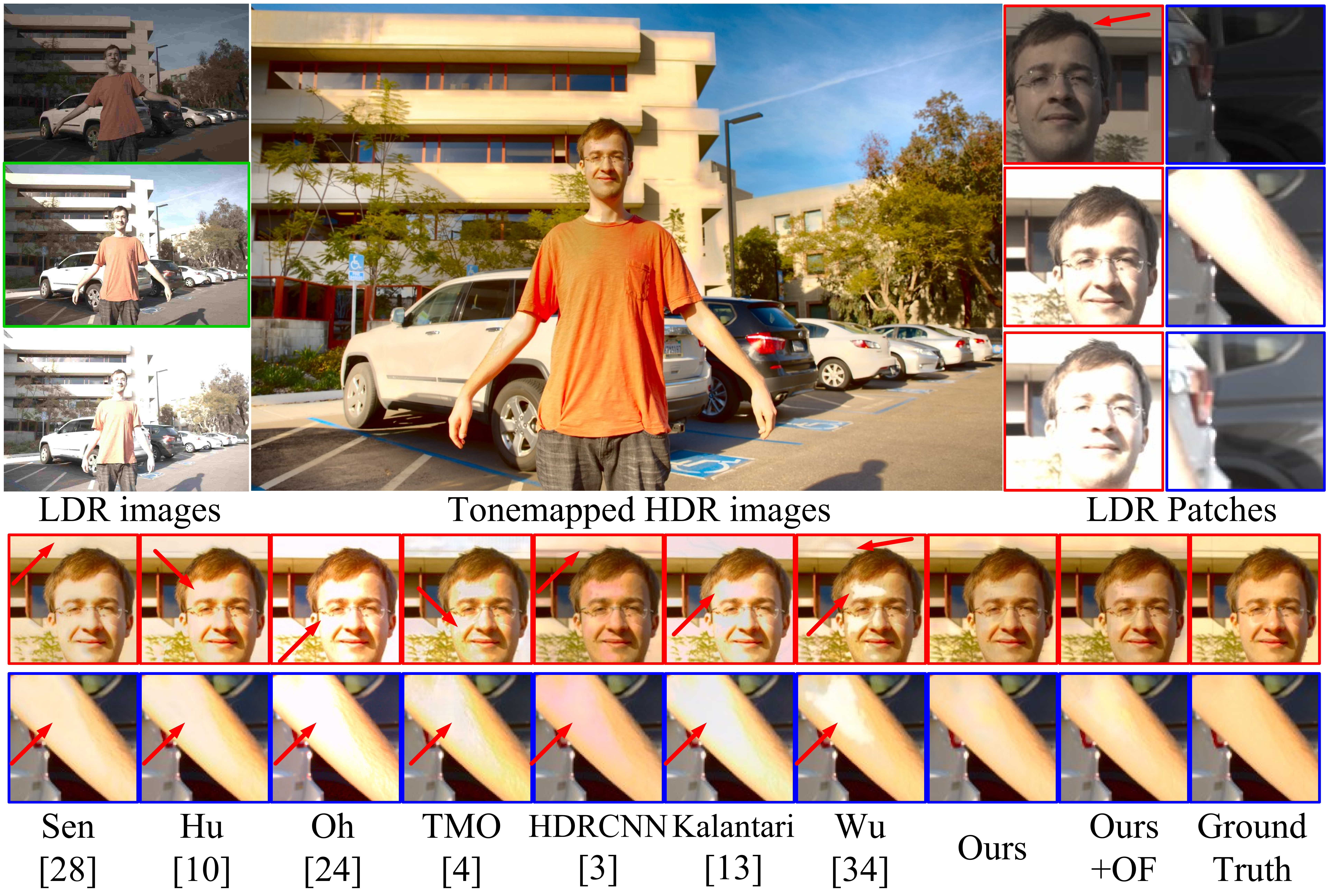}}
     \centerline{\begin{small}(b) Testing data (Parking) in \cite{Kalantari2017Deep}\end{small}}\medskip
\end{minipage}

\caption{Visual comparisons on the testing data from Kalantari \etal \cite{Kalantari2017Deep}. The top half part shows the input LDR images, LDR image patches, and the HDR image produced by the proposed method. We compare the zoomed-in local areas of the HDR images estimated by our methods and the compared methods. 
The propose network can produce a high-quality HDR image, especially saturated and object motions regions.
}
\label{Fig_compare_test}
\end{figure*}

\begin{figure*}[htbp]
\def \wid{8.5cm}
\def \Lwid{0.495}
\centering
\begin{minipage}[h]{\Lwid\linewidth}
      \centering
      \centerline{\includegraphics[width=\wid]{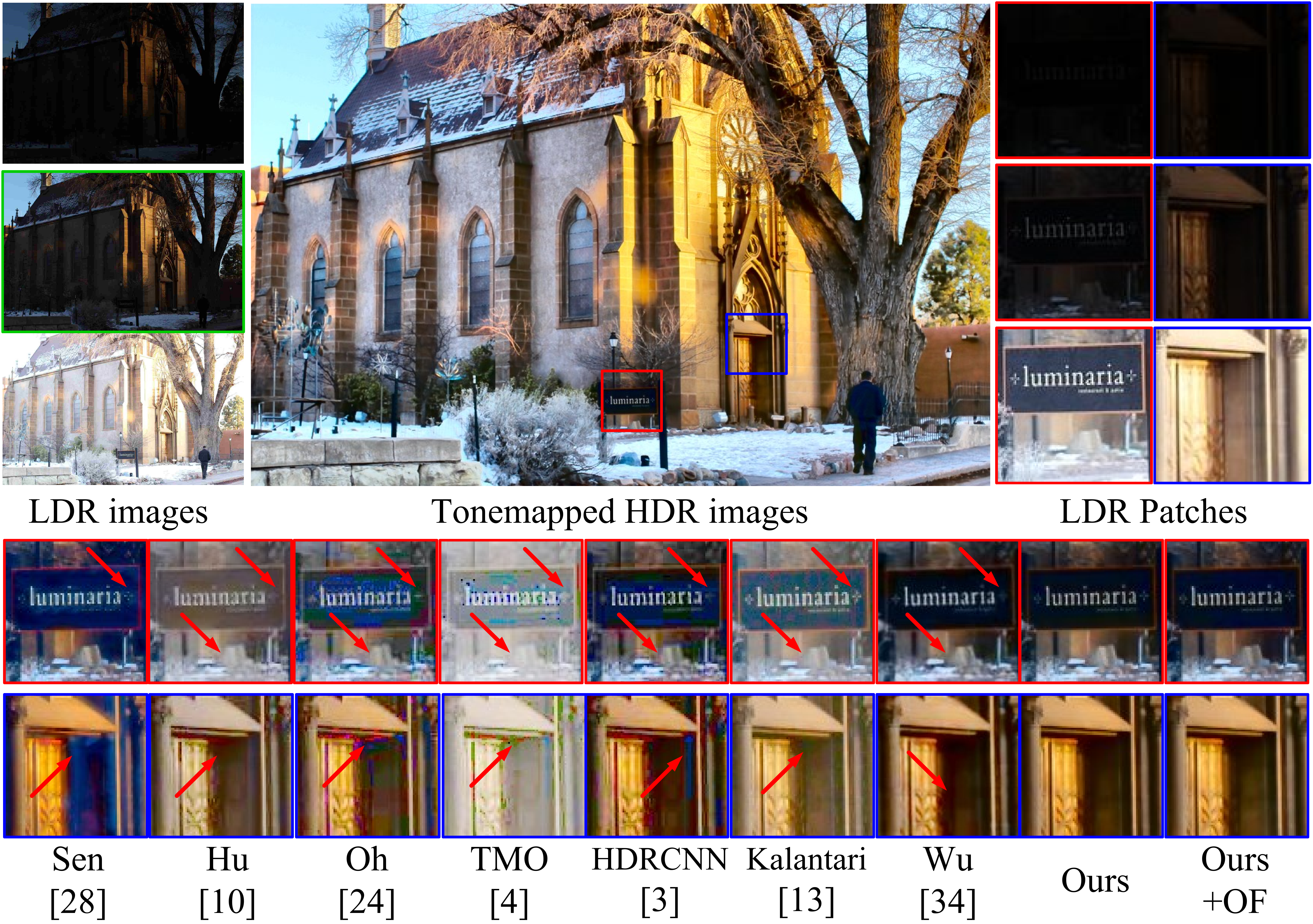}}
     \centerline{\begin{small}(a) Example from Sen \emph{et al.}'s dataset \cite{Sen2012}\end{small}}\medskip
\end{minipage}
     \hfill
\begin{minipage}[h]{\Lwid\linewidth}
      \centering
      \centerline{\includegraphics[width=\wid]{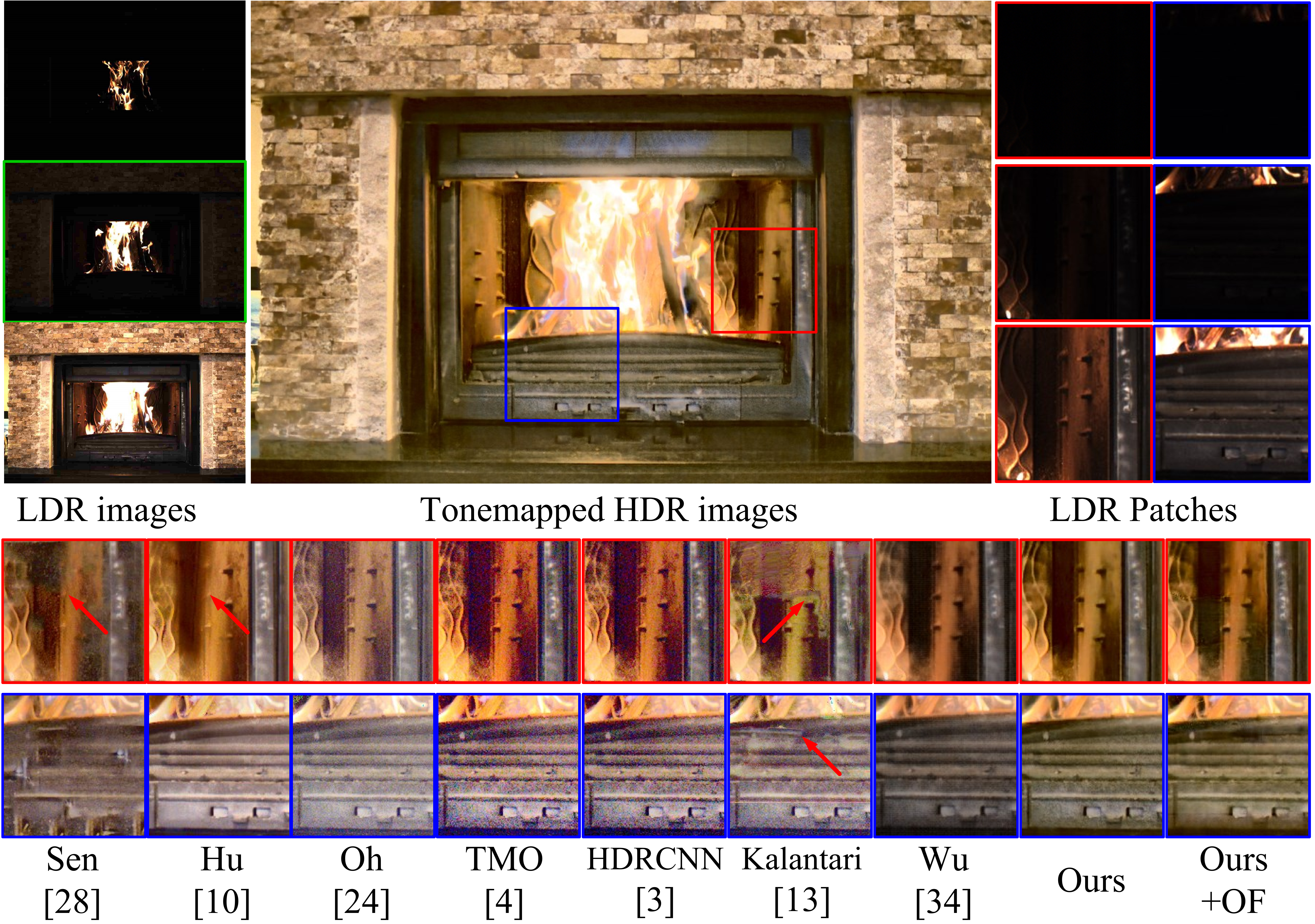}}
     \centerline{\begin{small}(b) Example from Tursun \emph{et al.}'s dataset \cite{Tursun2016data}\end{small}}\medskip
\end{minipage}

    \caption{Visual comparisons on the datasets without ground truth. The AHDRNet obtains results with sharper details and less artifacts.}
\label{Fig_compare_other}
\end{figure*}

\subsection{Comparison with the State-of-the-art Methods}
\par
We evaluate the proposed method and compare with previous state-of-the-art methods on a variety of datasets.
Specifically, we compare the proposed method with two patch-based methods \cite{Sen2012,Hu2013deghosting}, the method based on motion detection \cite{Oh2015pami}, the flow-based approach with DNN merger \cite{Kalantari2017Deep} and the DNN method without optic flow \cite{Wu_2018_ECCV}.
In addition, we compare with single frame HDR imaging methods \cite{Endo2017Deep,Eilertsen2017HDR}.
For all methods, we employed the codes provided by the authors.
The same training dataset and setting are used for deep learning methods.
Furthermore, we also apply the proposed AHDRNet with the input images aligned via estimated optical flow \cite{sun2014quantitative} (referred to as Ours+OF).

\def \blue{\textcolor{blue}}
\def \red{\textcolor{red}}

\begin{table}
\small
\begin{center}
\caption{Quantitative comparison of proposed network with several state-of-the-art methods. \red{Red} color indicates the best performance and \blue{blue} color indicates the second best result. }
  \label{tab:methods}
  \begin{tabular}{c|c|c|c|c}
  \hline
     &\footnotesize{PSNR-$\mu$} & \footnotesize{PSNR-M} & \footnotesize{PSNR-L} & \footnotesize{HDR-VDP-2}\\
     \hline
     Sen \cite{Sen2012}& 40.9453 &  30.5507 &  38.3147  &  55.7240\\
     Hu \cite{Hu2013deghosting}& 32.1872  &  25.5937 &   30.8395  &  55.2496\\
     Oh \cite{Oh2015pami} & 27.351 &  22.6311  &  27.1119  &  46.8259\\
     TMO \cite{Endo2017Deep}& 8.2123  &  21.4368  &   8.6846  &  44.3944\\
     HDRCNN \cite{Eilertsen2017HDR}&  14.0925  &  25.8217 &   13.1116  &  47.7399\\
     Kalantari \cite{Kalantari2017Deep}&  42.7423  & 32.0458 &  \textbf{\blue{41.2158}} &  60.5088\\
     Wu \cite{Wu_2018_ECCV} &41.6377 &31.0998 & 40.9082 & 60.4955 \\
     \hline
     Ours &  \textbf{\blue{43.6172 }}  & \textbf{\red{33.0429}}  & 41.0309 &  \textbf{\red{62.3044}}\\
     Ours + OF & \textbf{\red{43.9764}}  & \textbf{\blue{32.7785}}  & \textbf{\red{42.2883}} &\textbf{\blue{62.1296}}  \\
     \hline
\end{tabular}
\end{center}
\end{table}

\subsubsection{Evaluation on Kalantari \etal's \cite{Kalantari2017Deep} Dataset}
We compare our method with several state-of-the-art methods on the testing data of \cite{Kalantari2017Deep} (Figure \ref{Fig_compare_test} (a) and (b)), which contains some challenging samples with saturated background and foreground motions.
The patch-based methods (Sen \emph{et al.} \cite{Sen2012} and Hu \emph{et al.} \cite{Hu2013deghosting}) cannot find corresponding patches and produce artifacts (See the result in Figure \ref{Fig_compare_test} (a)).
The method of Oh \emph{et al.} \cite{Oh2015pami} cannot recover the details in the saturated areas.
Since the single image methods TMO \cite{Endo2017Deep} and HDRCNN \cite{Eilertsen2017HDR} only use the single reference image, they can avoid the ghosting artifacts, but are unable to reconstruct the sharp results and produces color distortion.
The method of Kalantari \emph{et al.} \cite{Kalantari2017Deep} products artifacts (See the red block in Figure \ref{Fig_compare_test} (a)), there have two main reasons: misalignment of optical flow and the limitation of their merging process.
Wu \emph{et al.}'s method \cite{Wu_2018_ECCV} generates over smooth results, and cannot completely remove the ghosting artifacts (See the red block in Figure \ref{Fig_compare_test} (a) and (b)).
Since our method uses the attention map (Figure \ref{Fig_attmap_small}) to select the useful regions and remove harmful components, it suppresses the ghosting artifacts and recovers the occluded or saturated details.
(See the blue block in Figure \ref{Fig_compare_test} (b)).
The proposed AHDRNet can produce high-quality results while taking the aligned images as inputs (See the results of Ours+OF) since the proposed attention module can also handle the artifacts caused by the error of alignment or optical flow estimation.

\par
As the ground truth is available for this testing set, we can conduct the quantitative evaluations and comparisons. Results are shown in Table \ref{tab:methods}. All the values are averaged over 15 testing images.
The proposed AHDRNet method produces better numerical performance than other methods.
The result is best in terms of PSNR-$\mu$ and PSNR-M, showing the effectiveness of the our model.
The proposed method (\ie Ours+OF) can produce slightly better or competitive results with the optical flow based alignment as preprocessing. 
With same alignment process, our method (Ours+OF) produces better results than \cite{Kalantari2017Deep}, which shows that our model can also help to handle the artifacts introduced by alignment error.

\subsubsection{Evaluation on the Datasets w/o Ground Truth}
We compare the proposed AHDRNet with other methods on Sen's \cite{Sen2012} and Tursun's \cite{Tursun2016data} datasets which do not have ground truth.
The results are shown in Figure \ref{Fig_compare_other} (a) and (b).
The patch-based Sen \emph{et al.}'s \cite{Sen2012} and Hu \emph{et al.}'s \cite{Hu2013deghosting} methods produce artifacts in complex motion regions (zoomed-in patches in Figure \ref{Fig_compare_other} (b)), these methods cannot find corresponding patches in the non-reference images.
As shown in Figure \ref{Fig_compare_other} (a) and (b), the single frame methods TMO \cite{Endo2017Deep} and HDRCNN \cite{Eilertsen2017HDR} prone to generate serious noise and color distortion in the under-exposed regions.
The method of Kalantari \emph{et al.} \cite{Kalantari2017Deep} introduction artifacts (Figure \ref{Fig_compare_other} (b)) due to the alignment error.
The results of Wu \emph{et al.}'s method \cite{Wu_2018_ECCV} miss details and have the obvious over smoothness in the results (Figure \ref{Fig_compare_other} (a) and (b)).
In comparison, our proposed AHDRNet produces appealing results where the geometry distortion, color artifacts, and noise are significantly reduced compared with existing methods.

\section{Conclusion}
\label{sec5}

The multiple exposure methods for HDR imaging can achieve high-quality outputs that better correspond to the dynamic range of the human visual system but has been limited in its application due to ghosting and saturating artifacts. 
The attention-based neural network we proposed overcomes these limitations. Most notably, it can generate high-quality HDR images even in the presence of large image motion and saturation. It thus offers the prospect of more extensive applications of HDR imaging. 

\newpage
{\small
\bibliographystyle{ieee_fullname}
\bibliography{egbib}
}

\end{document}